\documentclass{article}

\usepackage[applemac]{inputenc}
\usepackage[T1]{fontenc} 
\usepackage[english]{babel}
\usepackage{amsmath,amssymb}
\usepackage{graphicx}
\usepackage{lineno}

\title{Emergence of Concepts in DNNs?}

\author{Tim Räz\footnote{University of Bern, Institute of Philosophy,  L\"anggassstrasse 49a, 3012 Bern, Switzerland. E-mail: tim.raez@posteo.de}}

\begin{document}

\maketitle

\begin{abstract}
The present paper reviews and discusses work from computer science that proposes to identify concepts in internal representations (hidden layers) of DNNs. It is examined, first, how existing methods actually identify concepts that are supposedly represented in DNNs. Second, it is discussed how conceptual spaces -- sets of concepts in internal representations -- are shaped by a tradeoff between predictive accuracy and compression. These issues are critically examined by drawing on philosophy. While there is evidence that DNNs able to represent non-trivial inferential relations between concepts, our ability to identify concepts is severely limited.
\end{abstract}

\section*{Introduction}

There is a well-known story of how deep neural networks (DNNs) predict classes in an image classification task \cite{bengi2009,lecun2015}: In the hidden layers of DNNs, progressively abstract concepts are represented. Take a model that classifies animals such as cats, dogs, cows. According to the story, the model detects low-level concepts such as colors and textures in the first layers. In intermediate layers, the model detects higher-level concepts, such as body parts (eyes, ears), or complex textures (fur), by composing low-level concepts. In the final layer, the model detects animals by composing higher-level concepts. Importantly, these concepts are emergent, i.e., they are not hard-wired into the models and do not correspond to the labeled classes, but are acquired through the learning process.

The direct route to verify this story is to examine the internal representations (hidden layers) of DNNs and to identify concepts supposedly represented there. The main goal of the present paper is to review and discuss work from computer science that proposes to do this. Two issues with concepts in internal representations of DNNs are discussed. First, how do these methods actually identify concepts that are supposedly represented in a DNN? Second, how are conceptual spaces -- sets of concepts in internal representations -- shaped by the classes to be predicted and by the representational capacities of DNNs?  These questions are critically examined by drawing on philosophy.

\section*{Background}

\paragraph{Concepts}

Before concepts in internal representations are discussed, criteria for concept possession should be stated. There are various philosophical theories of concepts \cite{margo2021}. Here an undemanding theory, or explication, of concept possession is used, which does not assume that concept possession requires mental states or consciousness, or that concepts are abstract objects \cite{buckn2018}. Rather, concepts are taken to be associated with abilities. An important distinction is between the \emph{extension} and the \emph{meaning} (intension) of a concept -- the distinction goes back to Frege \cite{zalta2022}. The extension of a concept is the collection of entities falling under it. Here, DNNs are taken to ``recognize'' (partial) extensions through activation patterns in their hidden or output layers. Labeled input instances can be seen as (partial) extension of a concept. When humans label instances, they do this on the basis of prior knowledge about the instances -- the meaning of a concept. In DNNs, the possession of meaning encompasses the representation of some inferential relations, e.g.: a cat is an animal, has four legs, a head, fur (usually), and so on. It will be assumed that both the extension and the meaning of a concept are relevant. As we will see below, one of the main challenges in the context of DNNs is that concepts are primarily identified through (partial) extensions, which underdetermines meaning.

We will see evidence that DNNs learn non-trivial inferential relations between concepts. Internal representations arise as a function of the objective of predicting $n$ classes. The concepts we will focus on are not the predicted classes, but emerge as a function of the objective of learning to predict the classes. In particular, DNNs apparently learn concepts that are shared by several classes. To use the example of classifying animals, in order to classify cats and dogs, a DNN may learn concepts such as `fur', `head', `paw', `eye', and so on that are shared by cats and dogs. This issue has been explored for some time, e.g., DNNs supposedly exploit that many classes have shared features at least at a low level in order to improve generalization \cite{bengi2009}. If these findings can be confirmed and DNNs are in fact able to learn non-trivial inferential relations between concepts, the representation of concepts contains information about meaning rather than extensions.

It could be argued that the exercise of examining concepts in internal representations is superfluous, because the predictive successes of DNNs shows that they are able to automatically identify predictively salient concepts. If this were not the case, DNNs would not be able to generalize as well as they do. However, DNNs are not always successful; there are known failure modes such as adversarial examples \cite{szege2014}. Also, predictively successful models are not necessarily models that represent their target system adequately; this is true for scientific models as well as for DNNs, as philosophers know \cite{jebei2020,raez2020}.

\paragraph{Conceptual Spaces}

DNNs may be able to learn concepts that are relevant to predict more than one class. Other factors may shape how concepts are represented in DNNs as well. First, the predicted classes may not share certain concepts and be mutually exclusive (to some extent). In the example of animal classification, in order to classify cows, a useful concept to be learned by a DNN may be horns. This concept does not contribute positively to the classification of cats and dogs, because cats and dogs are not horned. Second, the concepts populating the internal representation take up some space in the internal representation, they are in competition for a finite amount of representational space. This competition may lead to compression and thus shape the internal representation of all concepts. Third, individual concepts may be compressed as well: if the representation of a concept contains predictively irrelevant details, this will lead to overfitting.

All these factors contribute to the formation of a conceptual space, the set of concepts in an internal representation of a DNN. Conceptual spaces are formed as a function of both the set of predicted classes and the representational capacity of the DNN. There have been some studies of how conceptual spaces are formed in DNNs, but less is known about this than about the emergence of individual concepts. Below we will see some evidence for compression due to competing concepts, and it will be argued that understanding conceptual spaces may be necessary to understand how individual concepts are represented in DNNs.

\paragraph{Limitations}

Not all aspects of the internal representation of a DNN have an interpretation in terms of concepts. An internal representation may not relate to any concept in that a) there may be a failure to represent, as in adversarial examples \cite{szege2014} or b) what is represented may not be accessible or comprehensible to humans and therefore not correspond to a concept \cite{boge2021}. Note that adversarial examples may also constitute predictively useful patterns, or artifacts \cite{buckn2020}. Cases a) and b) are examples of non-conceptual content of an internal representation. Understanding the scope and limits of non-conceptual content is important, but the following discussion will focus on the modes of representation of concepts that can be grasped by humans. The discussion will focus on post-hoc methods to extract concepts from trained models, excluding methods like concept whitening \cite{chen2020} or concept bottleneck \cite{koh2020} that modify the architecture of DNNs to enhance interpretability.

\section*{Identifying Emergent Concepts}

In this section, empirical work on the emergence of concepts in the internal representation of DNNs is reviewed. The focus is on concepts that are relevant to several of the predicted classes, because such concepts indicate non-trivial inferential relations.

\paragraph{Network Dissection}

Network dissection by Bau et al. \cite{bau2020} proposes to identify concepts associated with individual neurons in CNNs. Specifically, the emergence of object detectors in scene classifiers is examined. For example, the CNN learned the concept `airplane' in the process of classifying `airfield' and `hangar'. Concepts are identified by matching the region of the input that maximizes activation of a neuron with a region associated with a concept given by an image segmentation method. Many concepts were found to be important for the classification of multiple scenes.

Network dissection has several advantages: it is automatic, allows for a quantitative evaluation of similarity, and for visual inspection of image regions. A drawback is that the image segmentation method can only identify a fixed, limited set of concepts. If a concept is not included in this set, it cannot be identified. Therefore, network dissection falls prey to a version of the ``bad lot'' argument by van Fraassen \cite{vanfr1989, douve2021}: If we explain scientific evidence (here: region with high activation by a neuron) using the best hypothesis from a limited set (here: concepts from image segmentation) it is not clear that the best hypothesis is also true, because the true concept may simply not be in the scope of the image segmentation method. It could be thought that this problem can be overcome by visually inspecting the regions with high activation to identify the concept. Such a region, however, is only (part of) an extension of a concept, and it may be unclear what meaning is associated with that region -- a segment containing a plane can also be described as a tube with wings. This is a version of the so-called indeterminacy of reference described by Quine \cite{quine2013,micha2022}.

\paragraph{Feature Visualization}

Feature visualization by Olah et al. \cite{olah2017} proposes to identify concepts by constructing input instances that maximize the activation of neurons (or other parts) of CNNs. The method generates synthetic images that maximize activation of a neuron. Olah et al. note that direct, unregularized optimization can lead to degeneracies (akin to adversarial examples), and that different kinds of regularization have to be used to obtain natural-looking images. Feature visualization can be used to show that low-level concepts combine to higher level concepts, e.g., a car detector is assembled from features like windows, car body and wheels \cite{olah2020}.

Feature visualization has the advantage that one does not need to infer the meaning of a concept from a set of instances. Rather, it provides a single visualization (or few). Of course, one still needs to determine meaning from an instance. As Olah et al. acknowledge, while many visualizations have a rather clear semantic interpretation, some visualizations appear to have a mixed meaning (so-called polysemantic neurons, more on these below), and some visualizations have no discernible meaning at all. Thus, the indeterminacy of reference is an issue here as well. Furthermore, the visualizations depends on the choices made in optimization, the regularizations in particular, which may introduce artifacts. The use of optimization raises further concerns, for example, the method could get stuck in a local optimum. Optimization in DNNs is not very well understood from a theoretical point of view \cite{vidal2017,berne2021}, and the possibility of local optima makes the method susceptible to the ``bad lot'' argument. 

\paragraph{TCAV}

Testing with Concept Activation Vectors (TCAV) by Kim et al. \cite{kim2018} is a method to examine how strongly a user-defined concept is associated with a predicted class in a particular layer. Concepts are defined extensionally by a user through a set of input examples of that concept and a set of random counterexamples. The concept activation vector of a layer is the vector normal to the hyperplane that best separates the activations of examples and counterexamples. One can test how strong the association of this concept with a predicted class is by measuring how well its vector aligns with the vector of that class. Kim et al. claim that DNNs learn emerging concepts with considerable accuracy. Classifiers of low-level concepts (colors, shapes) achieve high accuracy in early layers, while more complex concepts (race, gender) achieve higher accuracy in later layers. Note that other researchers have explored the activation of layers with linear classifiers \cite{alain2018}.

The main advantage of TCAV is that is allows users to choose the concepts to be identified through customized sets of examples. TCAV thereby overcomes, to some extent, the philosophical problem of the indeterminacy of reference we encountered above: in principle, there is no limit on the number and variety of instances to define a concept extensionally. There are, of course, practical limitations. Also, the extensional definition of concepts limits the control on the meaning of the concept being defined. A further drawback of the method is its limitation to testing for linear information in the layers.

\paragraph{Non-local Representation of Concepts}

The above methods differ in how they identify concepts, but they also vary in \emph{where} they take concepts to be represented (in single neurons, layers, spread over several layers). It is known that concepts are not (only) represented by individual neurons, but have distributed representations. There is evidence that the representation of concepts is not limited to single layers. Yosinski et al. \cite{yosin2014} examined concept representations from the perspective of transfer learning. They found that feature representation in intermediate layers is distributed over consecutive layers: freezing only a portion of consecutive intermediate layers led to a worse performance than freezing all intermediate layers in question. This constitutes indirect evidence that the relevant concepts are distributed over these layers.

\section*{Conceptual Spaces}

In this section, it is discussed how conceptual spaces arise as a function of the predicted classes and of compression. The discussion is more speculative than the last section, because there is less empirical work on the global perspective of conceptual spaces. 

\paragraph{Polysemantic Neurons}

Some indirect, local evidence for competing concepts is provided by feature visualization, see above. If concepts are disjunctive and in competition for representational space, say, in a layer of a DNN, then one observable consequence may be that some concepts have imperfect representations and become mixed. This phenomenon has been observed by Goh et al. \cite{goh2021}. They find that while many neurons maximize activation for a single, identifiable concept, so-called \emph{polysemantic neurons} are composites of different, seemingly unrelated concepts, e.g., a neuron representing a mix of cats and cars. Goh et al. point out that one possible explanation of this sort of disjunctive neuron is that they could make ``concept packing more efficient''\cite{goh2021}. This idea is discussed in more detail by Olah et al. \cite{olah2020} as the \emph{superposition hypothesis}.

\paragraph{Completeness-aware Concept-based Explanations}

Completeness-aware concept-based explanations (CCE) proposed by Yeh et al. \cite{yeh2020} is a method geared towards discovering sets of concepts that are not only positively relevant to the predicted classes, but complete. A complete set is akin to a sufficient statistic, that is, a function of the input that retains all predictively relevant information \cite{casel2002}. CCE identifies concepts by partitioning linear directions in the activation space of a hidden layer, such that similar concepts are as close as possible, and dissimilar ones as distant as possible. The meaning of concepts is determined by inspecting input instances. This approach distinguishes itself by identifying complete sets of concepts as opposed to single concepts. However, it affords little control on whether the discovered concepts are meaningful.

\paragraph{Minimal Sufficient Statistics and the Information Bottleneck}

It is plausible that DNNs learn compressed, efficient representations of concepts, because the space to represent concepts is limited. From a statistical point of view, the layers of a DNN form a Markov chain, which means that information is lost and internal representations become more abstract in deeper layers \cite{alain2018,achil2018}. But what are the rules that guide how concepts are compressed? To understand these rules, a more global perspective on the representation of concepts is necessary. The CCE approach provides a global perspective in the form of a complete set of concepts, i.e., a sufficient statistic. However, in order to account for the idea of an \emph{efficient} representation of concepts, \emph{minimal} sufficient statistics (MSS) are relevant \cite{casel2002}. MSS are sufficient statistics that are as coarse as possible and thus provide the most efficient representation without losing predictive power. 

It has been argued that DNNs cannot learn minimal sufficient statistics \cite{schwa2017}. MSS only yields a useful degree of compression for a very particular kind of data distribution \cite{casel2002}, which is not given for most empirical datasets processed by DNNs. A helpful framework that generalizes MSS and can be applied to DNNs is the so-called \emph{Information Bottleneck} (IB) \cite{shami2010,schwa2017,raez2020a}. Tishby and collaborators propose that the IB explains how internal representations of DNNs arise as a tradeoff between predictive accuracy (sufficiency) and compression (minimality). Formally, the IB tradeoff is a constrained optimization problem, which yields a predictively optimal representation for a given level of compression (information loss). Tishby et al. have argued that layers of DNNs in fact approximate the optimum given by the IB tradeoff. Note that while the IB framework has been applied extensively, it has been contested whether it is an adequate account of how internal representations arise in DNNs \cite{saxe2018,geige2020}.

\paragraph{Visualizing Conceptual Spaces: Color Naming}

The IB framework provides a theoretical picture of how entire conceptual spaces emerge in DNNs. Unfortunately, there is a lack of work establishing relations between this theoretical picture and the actual representation of concepts in given DNNs. In order to illustrate what picture could emerge if conceptual spaces in DNNs were investigated, we will now consider an application of the IB framework to concepts in an empirical context.

\begin{figure}[h] 
   \centering
   \includegraphics[width=90mm]{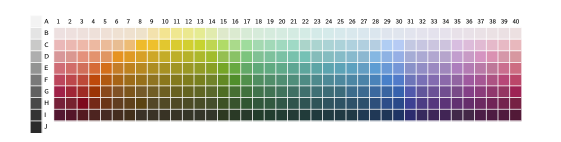} 
   \caption{The WCS palette. © by Zaslavsky et al. (2018).}
   \label{fig:wcs_palette}
\end{figure}

Zaslavsky et al. \cite{zasla2018a} use the IB framework to explain how color naming systems arise as a result of efficiency. Different natural languages use different systems to name colors. Based on a standard representation of colors (the WCS stimulus palette, see Figure \ref{fig:wcs_palette}), one can determine how speakers of different languages name the color chips on this palette.

\begin{figure}[h] 
   \centering
   \includegraphics[width=120mm]{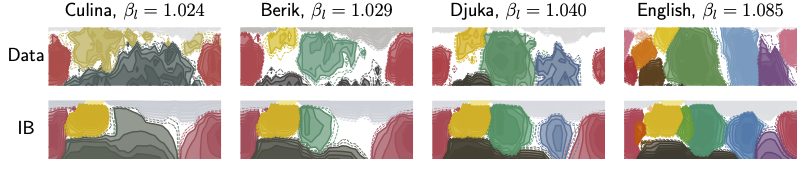} 
   \caption{Color naming of four different languages, determined empirically (top row) and theoretically from the IB (bottom row). The parameter $\beta_l$ controls the degree of compression in the IB tradeoff. © by Zaslavsky et al. (2018).}
   \label{fig:color_naming}
\end{figure}

This yields different (soft) partitions of the color space, corresponding to the color systems of these languages, cf. Figure \ref{fig:color_naming}, top row.

Zaslavsky et al. propose a theoretical explanation of how these different color naming systems of different languages arise. They argue that the different empirical partitions (Figure \ref{fig:color_naming}, top row) match closely with partitions that are derived from the IB framework (Figure \ref{fig:color_naming}, bottom row). The main difference between languages is the number of color concepts they use. A language with more color concepts yields a more fine-grained partition, a language with less colors a more coarse-grained partition. The partitions derived from the IB framework are determined to a large degree by the tradeoff between accuracy and compression, controlled by the parameter $\beta_l$, which yields different numbers of concepts (the theoretical predictions also depend on the so-called least informative prior). The close fit between theoretical and empirical partitions suggests that color naming systems in different languages have evolved to communicate accurately about colors at a given level of compression, and that the level of compression is due to the different communicative needs of the societies using the languages.

How is this related to conceptual spaces in DNNs? To spell out the analogy, the naming systems correspond to internal representations, e.g., partitions of activation patterns in a hidden layer. The cells of the partition (colors) correspond to clusters of activation patterns with a meaning (concepts). The degree of compression is measured by the number of concepts, and the number of concepts is determined by communicative need in the case of color naming systems, and the predicted classes and representational capacity in the case of DNNs. The analogy is substantive to the extent that both color spaces and representations in DNNs are driven by the IB objective.

The analogy allows us to get a sense of how sets of concepts may emerge holistically in DNNs, that is, as a function of predicting classes while having a limited representational capacity. If we compare the different partitions in Figure \ref{fig:color_naming}, we can see that as the number of colors changes, the entire partition changes. This illustrates how the representation of concepts depends on the representational capacity. Note that polysemantic neurons would indicate an internal representation that is too compressed for the concepts to be represented, such that two concepts (colors) merge.

The analogy has its limits. For one, color naming is special because colors are disjunctive, which need not be the case for other concepts. Also, the representation of colors is non-hierarchical, in contrast to complex representations in DNNs. Note that the conceptual spaces of other kinds of objects have been investigated, but they do not allow for similarly striking visualizations \cite{zasla2019}. An important open question about compressed representations concerns the mechanism by which compression is achieved. It is unclear whether compression is due to limited representational space, because many successful DNNs are overparametrized, as witnessed by the double-descent risk curve \cite{belki2019,berne2021}. Compression could also be an effect of randomness induced by stochastic gradient descent \cite{schwa2017}.

\section*{Discussion}

\paragraph{Robust Identification of Concepts}

Network dissection and TCAV identify concepts via partial extensions, which leads to problems because partial extensions underdetermine the meaning of concepts. Feature visualization relies on optimization, which raises other issues. These problems can be seen as in-principle, philosophical obstacles to identifying concepts in DNNs. From a more pragmatic perspective, the individual weaknesses of these methods could be overcome to some extent by combining them and performing what is known as \emph{robustness analysis}. Robustness analysis, first proposed in population biology, determines whether different, imperfect methods arrive at the same prediction to increase reliability, under the slogan: truth is at the intersection of independent lies \cite{levin1966,wimsa2012,wimsa2012a,knuut2011}. One could apply methods like TCAV and feature visualization to the same model. If different methods identify the same concept independently, this should raise our confidence that the methods are somewhat reliable. Robustness analysis is limited in that it will not yield an absolute confirmation of concepts \cite{orzac1993} -- it is only as good as the set of methods in combination -- but it is better than using only one method. A combination of different methods contributing to interpretability has been proposed and explored \cite{olah2018,kim2018}.

\paragraph{Testing Methods with Synthetic Data}

The methods for identifying concepts considered here are limited to extracting local or linear information. It would be desirable to extend the scope of the methods to encompass the identification of concepts with distributed and non-linear representations. However, this will be hard to carry out by sticking to the extensional paradigm of identifying concepts. Defining concepts with sets of instances only allows for limited control on the meaning of concepts. One possibility to gain more control on meaning would be to create synthetic datasets in which not only the predicted classes (animals) are labeled, but also intermediate concepts (body parts, textures, etc.), which may re-emerge in internal representations of DNNs -- interpretable datasets, so to speak. This approach has been proposed in the context of interpretable architectures \cite{koh2020}. However, synthetic datasets could also be used to test methods for non-interpretable architectures, such as TCAV or feature visualization. In the context of physical modeling, the use of simulation data has led to some progress in developing DNN emulators for which emerging, high-level properties (e.g. energy conservation) can be checked \cite{Gentine-et-al2018,beucler2019achieving,Couvreux-et-al2020}. One of the main challenges of this approach would be to come up with a principled labeling system for the intermediate concepts.

\paragraph{The Need for Conceptual Spaces}

We have discussed the identification of both concepts and conceptual spaces. It could be asked whether both are really necessary, because once we have identified all the concepts in an internal representation, we have arguably also identified the conceptual space. This argument presupposes that the identification of individual concepts in internal representations is reliable and leads to a neat partition of an internal representation. However, this presupposition is not realized in practice. Methods to identify particular concepts, and also sets of concepts, are not (yet) reliable. Also, conceptual spaces may contain elements like polysemantic neurons, as well as artifacts, which do not have neat conceptual counterparts. Understanding how entire conceptual spaces are formed is an additional path to understanding how individual concepts are formed. Bottom-up methods, which identify concepts, and top-down methods, which examine entire conceptual spaces, should not be seen as competing, but as complementary ways of triangulating concepts in internal representations, ultimately making the triangulation more reliable.

\section*{Conclusion}

There is evidence that DNNs able to represent non-trivial inferential relations between predicted classes and emergent concepts that are represented internally. This indicates that DNNs may be able to acquire information that is not purely extensional. However, our ability to identify emergent concepts in the first place is severely limited, because existing methods rely on limited extensions of concepts, which makes them susceptible to philosophical problems such as the indeterminacy of reference and the bad lot argument. These limitations should give us pause, given that we have used an undemanding theory of concepts. Finally, the problem of understanding how entire sets of concepts arise holistically in internal representations through tradeoffs between predictive accuracy and compression is underexplored. Novel methods to identify concepts as well as conceptual spaces are urgently needed.

\end{document}